\documentclass{article}

\usepackage{graphicx}

\usepackage[utf8]{inputenc} 
\usepackage[T1]{fontenc}    
\usepackage{hyperref}       
\usepackage{url}            
\usepackage{booktabs}       
\usepackage{amsfonts}       
\usepackage{nicefrac}       
\usepackage{microtype}      
\usepackage{graphicx}
\usepackage{subcaption}
\usepackage{color}
\usepackage{amsmath}
\usepackage{amssymb}
\usepackage{multirow}

\usepackage{hyperref}

\usepackage[accepted]{arxiv2021}
\usepackage{array}
\newcolumntype{M}[1]{>{\centering\arraybackslash}p{#1}}

\icmltitlerunning{Unsupervised Object-Based Transition Models For Embodied Agents in 3D Partially Observable Environments}

\begin{document}

\twocolumn[
\icmltitle{Unsupervised Object-Based Transition Models for 3D Partially Observable Environments}

\begin{icmlauthorlist}
\icmlauthor{Antonia Creswell}{dm}
\icmlauthor{Rishabh Kabra}{dm}
\icmlauthor{Chris Burgess}{dm}
\icmlauthor{Murray Shanahan}{dm}
\end{icmlauthorlist}

\icmlaffiliation{dm}{DeepMind, London}

\icmlcorrespondingauthor{Antonia Creswell}{tonicreswell@google.com}

\icmlkeywords{Objects, Transition models}

\vskip 0.3in
]

\printAffiliationsAndNotice{}

\begin{abstract}
We present a slot-wise, object-based transition model that decomposes a scene into objects, aligns them (with respect to a slot-wise object memory) to maintain a consistent order across time, and predicts how those objects evolve over successive frames. The model is trained end-to-end without supervision using losses at the level of the object-structured representation rather than pixels. Thanks to its alignment module, the model deals properly with two issues that are not handled satisfactorily by other transition models, namely object persistence and object identity. We show that the combination of an object-level loss and correct object alignment over time enables the model to outperform a state-of-the-art baseline, and allows it to deal well with object occlusion and re-appearance in partially observable environments.
\end{abstract}

\begin{figure*}[h!]
    \centering
    \includegraphics[width=0.8\textwidth]{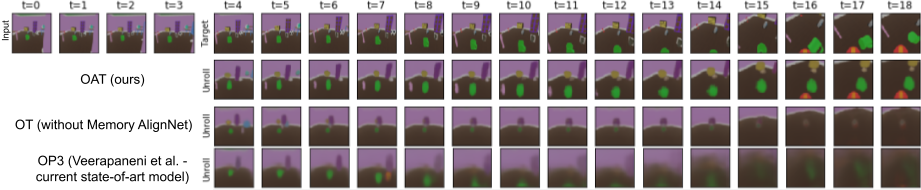}
    \caption{\textbf{Comparing our model to baselines.} Each model was trained with four input steps and to unroll for six steps, here we unroll for 15 steps. Our model, OAT, performs significantly better than OT (our model without the AlignNet) and current state-of-the-art model, OP3 \cite{veerapaneni2019entity}. See Figure \ref{fig:op3_rollouts} in the Appendix for additional OP3 roll-outs.}
    \label{fig:MAT_vs_baselines}
\end{figure*}

\section{Introduction}

In spite of their well-documented ability to learn complex tasks, today’s deep reinforcement learning agents are still far from matching humans at out-of-distribution generalisation or few-shot transfer \cite{garnelo2016towards, lake2017building, marcus2019rebooting}. Two architectural features commonly proposed to remedy this are (1) transition models that enable the agent to internally explore paths through state space that it has never experienced \cite{silver2016mastering, racaniere2017imagination, ha2018world}, and (2) compositionally structured representations that enable the agent to represent meaningful states that it has never encountered \cite{garnelo2019reconciling}. These two features are not exclusive; transition models that operate on compositionally structured representations are a potent combination, and these are the subject of the present paper. Specifically, our focus is on transition models that operate at the level of {\em objects}, which are the most obvious candidates for the structural elements of representations likely to be useful for artificial agents inhabiting 3D worlds such as our own \cite{shanahan2020artificial}.

While there has been progress in object-based transition models \cite{watters2019cobra, veerapaneni2019entity, weis2020unmasking, kosiorek2018sequential}, current models do not deal satisfactorily with {\em object persistence} (the concept that objects typically continue to exist when they are no longer perceptible \cite{Pylyshyn1989TheRO}) or with {\em object identity} (the concept that a token object at one time-step is the same token object at a later time-step \citet{baillargeon1985object}). As we show, transition models that neglect object persistence tend to perform badly in complex, partially observable environments, while models that neglect object identity are unable to integrate information about a single object (and its interactions) over time in a way that generalises to future time-steps. By incorporating a module for aligning objects across time using a slot-based memory, our model handles both these concepts, and exhibits better performance as a result.

An important feature of our transition model is that it makes predictions and computes losses in a representation space divided into objects. This contrasts with existing models that make predictions in an unstructured representation space \cite{hafner2019dream, ha2018world}. Making predictions and computing losses in an object-structured representation space facilitates learning, not only because the representation space is lower dimensional than pixel space, but also because the model can exploit the fact that dynamics tend to apply to objects as a whole, which simplifies learning. However, to compute prediction losses directly over distinct object representations (rather than first mapping predictions back to pixels \cite{watters2019cobra, veerapaneni2019entity}), the objects in a predicted representation must be matched with those in the target representation, which again requires a proper treatment of object persistence and identity. The result is a model that can roll-out accurate predictions for significantly more steps than those seen during training, outperforming state-of-the-art for comparable models.

To achieve this, our model, Objects-Align-Transition (OAT), combines (1) a {\em scene decomposition and representation module}, MONet \cite{burgess2019monet}, that transforms a raw image into a slot-wise object-based representation, (2) an {\em alignment module} which, with the aid of a slot-wise memory, ensures that each object is represented in the same slot across time, even if it has temporarily disappeared from view, and (3) a slot-wise {\em transition model} that operates on the object representations to predict future states. All three components are differentiable, and the whole model is trained end-to-end without supervision. We evaluate the model on two sequential image datasets. The first is collected from a pre-trained agent moving in a simulated 3D environment, where we show that our model outperforms the current state-of-the-art object-based transition model (OP3 \citet{veerapaneni2019entity}). The second dataset is collected from a real robot arm interacting with various physical objects, where we demonstrate accurate roll-outs over significantly longer periods than used in training. An ablation study shows that the model's success depends on the combination of correct object alignment across time and the use of loss over object-level representations instead of over pixels.

\section{Our Model: Objects-Align-Transition} \label{sec:model}

\begin{figure}[h]
    \centering
    \includegraphics[width=0.4\textwidth]{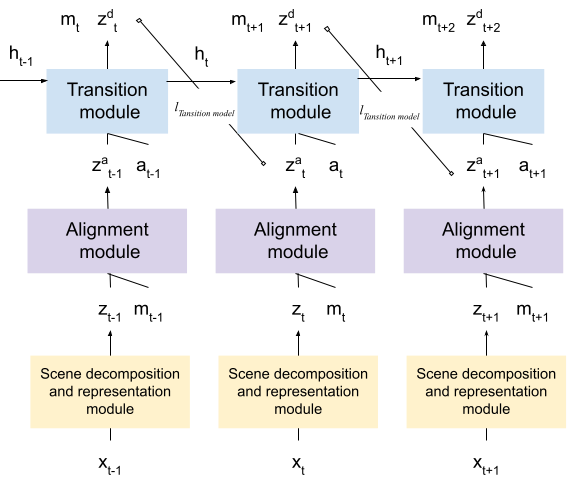}
    \caption{\textbf{Encoding steps.} The \textit{scene decomposition and representation module} extracts $K$ object representations, $z_t \in \Re^{K \times F}$, from images, $x_t \in \Re^{W \times H \times 3}$. These are aligned using the \textit{alignment module} to obtain, $z^a_t \in \Re^{M \times F}$. Aligned objects are fed to the slot-wise \textit{transition model} with the action, $a_t$, and hidden state, $h_t (= h^a_t) \in \Re^{M \times H}$, to predict the object representations at the next time-step, $z^d_t \in \Re^{M \times F}$ as well as the updated memory, $m_{t+1} \in \Re^{M \times F}$. On this figure we also demonstrate how the transition model loss is computed between $z^d_t$ and $z^a_t$.} \label{fig:enc}
\end{figure}

Our model, Objects-Align-Transition (OAT), combines a scene decomposition and representation module, in this case MONet, with a slot-wise transition module, via an alignment module, which ensures that each slot in the transition model receives objects corresponding to the same token object (or identity) across time. The whole model is trained end-to-end. We now provide details of each of these modules (see Figure \ref{fig:enc}).

\subsection{Scene decomposition and representation module}
The input to OAT is a sequence of RGB images, $x \in [0, 1]^{T \times W \times H \times 3}$, with $T$ time-steps, width, $W$, and height, $H$. We leave out the batch dimensions for simplicity. Each image in the sequence, $x_t$, is passed through a scene decomposition and representation module\footnote{We tried some variants where $z_t$ depended on $z_{t-1}$. However these did not yield significantly beneficial results.}, in this case MONet, to obtain a slot-wise object representation, $[z_{t, k}]_{1:K}$, occupying $K$ slots, where $z_{t, k} \in \Re ^F$ is an object representation vector with $F$ features. 

MONet consists of an attention module (a U-net \citet{ronneberger2015u}) --- which predicts object segmentation masks, $\mu_{t, k} \in [0, 1]^{T \times W \times H \times 1}$, for each slot --- and a slot-wise VAE. The encoder of the slot-wise VAE is fed the predicted object segmentation masks and the input image, $x_t$, and outputs object representations, $z_t \in \Re^{K \times F}$. The decoder reconstructs the masks, $\tilde{\mu}_{k, t}$, and each object's pixels, $\tilde{x}_{t, k}$.

\begin{equation} \label{eqn:monet}
\small
    \begin{split}
         \mu_{k, t} &= \text{Attention\_Network} (x_t)\\
         z_{k,t} &= \text{MONet\_encoder}(x_t, \mu_{k, t}) \\
        \tilde{\mu}_{k, t}, \tilde{x}_{t, k} &= \text{MONet\_decoder}(z_k)\\
    \end{split}
\end{equation}

It may be tempting to feed the slot-wise object representations, $[z_{t, k}]_{k=1:K}$, at time, $t$, directly to a slot-wise transition model. However, MONet's representations \cite{burgess2019monet} are \textit{not stable} across time, meaning that a specific object may appear in different slots across time (\citet{weis2020unmasking} and see Figure \ref{fig:playroom_monet}). This leads to two major problems when training slot-wise transition models on slot-wise object-based representations. Firstly, it is difficult to compute losses between predicted and target object slots. Computing a slot-wise \textit{object-level loss} requires us to know how the previous object representations (and thus the predicted object representations\footnote{because we compute $z_{t+1} = z_t + \Delta_t$}) correspond with the target object representations. Secondly, if object representations do not occupy consistent slots it makes it harder to integrate information about a single object (and its interactions with other objects) across time \cite{weis2020unmasking}, and makes it harder to predict the reappearance of that object (as show in Section \ref{sec:matters}). To address this slot stability issue we use an alignment module.

\subsection{Alignment module} \label{sec:align}

The alignment module plays two key roles. The first is to align objects across time, enabling us to compute slot-wise object-level losses for training. The second is to learn a slot-wise memory that encodes the history of each slot across time, allowing our transition model to operate effectively in partially observable environments by keeping tack of objects as they go in and out of view. This is especially important for embodied agents that take actions in 3D environments, where the agent's looking around frequently causes objects to move in and out of its field-of-view.

Here we use the Memory AlignNet \cite{creswell2020alignnet}, which takes a slot-wise memory, $m_t \in \Re^{M \times F}$, with $M \geq K$ slots and the (stacked) output of the scene decomposition and representation module, $z_t \in \Re^{K \times F}$, returning the aligned object representations, $z^a_t \in \Re^{M \times F}$. To perform alignment, the Memory AlignNet predicts an adjacency matrix, $A_t \in \Re^{M \times K}$, that aligns objects, $z_t$, with the memory, $m_t$. This adjacency matrix allows us to compute a \textit{soft} alignment $z^{a, soft} = A_t z_t$ or a hard alignment, $z^a \triangleq z^{a, hard} = \text{Hungarian}(A_t)z_t$. We will refer to $z^{a, hard}_t$ as $z^a_t$ throughout the rest of the paper. Hungarian$(\cdot)$ denotes the application of the Hungarian algorithm, a non-differentiable algorithm which computes a permutation matrix given an adjacency matrix\footnote{Since the adjacency matrix is not square we append extra columns to it and we exclude empty slots from the assignment process. Whether a slot is empty can be computed by looking at the masks output by MONet. If the sum of pixels in a mask is less than some threshold then we consider that slot to be empty}. The soft version of the alignment is used to train the AlignNet while the hard version of the alignment is passed to the slot-wise transition model, which has the same number of slots, $M$, as the memory. 

The hard aligned objects may also be used to update the memory using a recurrent slot-wise model. For simplicity we use our transition model, $\mathcal{T}_\theta(\cdot)$, to predict deltas for both the object representations, $z^a_t$, and the memory, $m_t$. We will detail this in the next section.

\subsection{Slot-wise Transition module}

The transition model operates in both an encoding and an unrolling phase. During the encoding phase the transition model is fed aligned, observed object representations, $z^a_t$, and actions, $a_t$, to predict aligned object representations at the next time-step, $z^d_{t+1} \in \Re^{M \times F}$. During the unroll phase, the transition model is fed the predictions, $z^d_t$, and actions, $a_t$, from the current time-step to predict the object representations, $z^d_{t+1}$, at the next time-step.

More concretely in the \textbf{encoding} steps, our transition model, $\mathcal{T}_\theta(\cdot)$, takes aligned object representations, $z^a_t$, and a hidden state, $h^a_t \in \Re^{M \times H}$, and predicts deltas for both the object representations, $z^a_t$, and memory, $m_t$. Concretely, the output of our transition model is given by, $[\Delta^a_t, \Delta^m_t], h^a_{t+1} = \mathcal{T}_\theta(z^a_t, a_t, h^a_t)$. The memory and the object representations are then updated as follows: $m_{t+1} = m_t + \Delta^m_t$ and $z^d_{t+1} = z^a_t + \Delta^a_t$. In the \textbf{unroll} steps, next step predictions are given by $z^d_{t+1} = z^d_t + \Delta^d_t$ where $[\Delta^d_t, \Delta^m_t], h^a_{t+1} = \mathcal{T}_\theta(z^d_t, a_t, h^a_t)$. By using the transition model to predict deltas, $z^d_t$, is aligned by default and does not need to be re-aligned when unrolling the model.

A key feature of our transition model is that weights are shared across object slots and can be instantiated in many different ways. One simple option is to use a \textit{SlotLSTM}; an LSTM applied independently to each slot, sharing weights between slots. An alternative instantiation which we found to work well is to first apply a transformer \cite{wang2018non, vaswani2017attention} and then the SlotLSTM. This allows the model to capture interactions with other objects (via the transformer) and integrate that information over time (via the SlotLSTM).

\subsection{Training}

When training OAT we jointly learn the parameters of the scene decomposition and representation module, the alignment module and the transition module. Each module has its own losses that contribute to downstream gradients and updates. For example, both the transition model and the alignment module losses can influence the scene decomposition and representation module's updates. Let us define the losses for each module.

\textbf{Scene decomposition and representation module losses (MONet)}. We train MONet using standard MONet losses, $l_{\text{MONet}}$, see Equation 3 in \citet{burgess2019monet}. The first term in the MONet loss is a scene reconstruction term. This is a spatial mixture loss parameterised by $\sigma_{bg}$ and $\sigma_{fg}$, which are the standard deviation used for each slot's component likelihood (for the first slot and remaining slots, respectively) that go into the mixture loss. The remaining terms are regularisation terms that 1) induce a latent information bottleneck needed for good representation learning (scaled by $\beta$, the latent KL loss weight), and 2) ensure that predicted and target masks are similar (scaled by $\gamma$, the mask KL loss weight).

\textbf{Alignment module losses (Memory AlignNet).} The memory AlignNet consists of a reconstruction loss between the softly aligned object representations, $z^{a, soft}_t$ and the output of the transition model, $z^d_t$ for the corresponding time-step. There are also regularisation losses including an entropy loss, $\mathbb{H}(\cdot)$, on the adjacency matrix, $A_t$, which encourages values towards zero and one; and a loss that penalises columns that sum to more that one, avoiding the case where multiple objects are assigned to the same memory slot. The AlignNet loss, $l_{\text{AlignNet}}$ is defined by Equation \ref{eqn:align}, where $T$ is the total number of encoding and unroll steps. We use $\psi=0.01$ for all experiments presented in this paper.

\begin{equation} \label{eqn:align}
\small
    \begin{split}
        \small
         l_{\text{AlignNet}} & =  \sum_{t=1}^T ||z^d_t - z^{a, soft}_t||_2^2 + \psi \mathbb{H}(A_t) \\
        & + \sum_{j=1}^M\max(0, (\sum_{k=1}^K A_{t, k, j} - 1)) \\
    \end{split}
\end{equation}

\textbf{Slot-wise transition module losses}. The slot-wise transition model is trained with a reconstruction loss between the outputs of the transition model, $z^d_t$, and the aligned observations for the same time-step, $z^a_t$. The slot-wise transition model loss, $l_{\text{Transition model}}$ is defined by Equations (\ref{eqn:transit}). Importantly, we compute a loss directly between object representations without the need for decoding them to obtain pixels.

\begin{equation} \label{eqn:transit}
\small
    \begin{split}
        & l_{\text{Transition model}}  = \sum_{t=1}^T||z^d_t - z^a_t||_2^2\\
    \end{split}
\end{equation}

OAT is trained end-to-end to minimise $l_{\text{MONet}} + l_{\text{AlignNet}} + \zeta l_{\text{Transition model}}$, using $\zeta=10$ for all experiments presented in this paper.

\section{Related Work} \label{sec:related_work}

Our work addresses the challenging topic of learning object-based transition models in 3D partially observable environments without supervision. We identify two main components missing from current object-based transition models, an understanding of object persistence and identity meaning that current models are not able to perform well in \textbf{partially observable environments} and that models are often trained without \textbf{semantically meaningful losses.}

Firstly, we acknowledge previous work on temporally extend scene decomposition and representation models \cite{weis2020unmasking, he2018tracking}, typically used for video representation learning, object-based transition models \cite{kosiorek2018sequential, hsieh2018learning} and action conditional object-based transition models such as OP3 \cite{veerapaneni2019entity} and C-SWM \cite{kipf2019contrastive}. We consider OP3 \cite{veerapaneni2019entity} to be the current state-of-the-art object-based transition model since C-SWM requires privileged information about exactly which object an action was applied to, while OP3 (like our model) simply requires the action taken by the agent.

\subsection{Current models are not designed for partially observable environments.}

The world that we operate in, and which we intend our agent's to operate in is partially observable, a severe limitation of existing object-based transition models \cite{veerapaneni2019entity, kipf2019contrastive} and temporally extended object scene decomposition and representation models \cite{weis2020unmasking, greff2016tagger, van2018relational, he2018tracking} is their inability to cope with partially observable environments. A promising approach proposed by \citet{he2018tracking} uses an external memory for object tracking\footnote{\citet{he2018tracking} do not propose a transition model.}, their mechanism is different to that in the Memory AlignNet \cite{creswell2020alignnet} which we use to perform slot alignment in OAT. \citet{he2018tracking} train their model, TBA, for reconstruction while the Memory AlignNet incorporates dynamics and is trained using prediction, meaning that the Memory AlignNet can resolve ambiguities using dynamics\footnote{For example when two visually similar objects collide with one another, the Memory AlignNet can use dynamics to resolve which object is which after the collision.}, while TBA cannot. Additionally, TBA can only cope with static backgrounds and so it not applicable here.

Further, a scene may contain $M$ objects, but at any point in time an agent may only see $K \leq M$ objects. Current \cite{veerapaneni2019entity, watters2019cobra, hsieh2018learning} models are restricted to either setting $K=M$ and extracting more entities per time-step than are needed, which may be computationally expensive, or setting $K<M$ and not accounting for some objects. Our model is able to achieve the best of both by incorporating a persistent memory with $M$ slots; the scene decomposition model extracts $K < M$ objects at each time-step, but still makes predictions over all $M$ persistent object slots.

\subsection{Current models are not trained using semantically meaningful losses.}
Another problem with current approaches to learning object-based transition models is how they are trained. In most scenes the background accounts for most of the pixels while the objects account for only a small fraction of them. So while it may be tempting to train transition and video-representation models using a pixel-level loss \cite{watters2019cobra, weis2020unmasking, veerapaneni2019entity, kosiorek2018sequential, hsieh2018learning, he2018tracking}, we show that is is preferable to compute losses directly between predicted and target objects (see  Section \ref{sec:matters} and Figure \ref{fig:MAT_vs_baselines}). Further, pixel-level losses require the object representations to be decoded into an image \cite{watters2019cobra, weis2020unmasking, veerapaneni2019entity} which is often computationally expensive.

An alternative way to compute object-level losses is to compute a minimum assignment loss \cite{reyes2019learning}, employing the Hungarian algorithm, between predicted and target object representations. However, doing so can be problematic because you need to first compute a similarity matrix on which to apply the Hungarian algorithm. We show in Section \ref{sec:matters} that computing a loss using the Hungarian algorithm\footnote{for the similarity matrix we use the $L_2$ loss between all object pairs.}, leads to poor generalisation in transition models when performing longer unrolls than those seen during training.

Furthermore, at the start of training comparing predictions with targets to compute a loss may not be meaningful since predictions will start off as essentially random. In our approach, we align the \emph{inputs} with the targets, and condition the slot-wise prediction on the slot-wise history. This reinforces slot stability and allows us to accurately predict changes, $\Delta^a_t$ in each object representation.

Interestingly Lowe et al. \cite{lowe2020learning} use Deep Sets \cite{zaheer2017deep}  to encode predicted and target object sets and use a contrastive loss between encodings. Their approach has only been demonstrated for very simple datasets. C-SWM \cite{kipf2019contrastive} do compute losses in their representation space, however they avoided the object correspondence problem because they extracted representations spatially, keeping the spatial ordering and using data where object movement was limited. Their approach is unlikely to scale well to partially observable environments with significant movement of objects across the field of view, especially when being trained to predict multiple steps in to the future.

OP3 \cite{veerapaneni2019entity} and other models \cite{weis2020unmasking} attempt to induce a weak, implicit object alignment by conditioning predicted object representations on those from the previous time-step. This does not guarantee alignment (or slots with consistent identity) over time, especially in partially observable environments and they do not use the alignment for computing losses.

Finally, we note that while there are existing models that learn to predict future states, given actions, directly from pixels \cite{hafner2019dream, ha2018world} we have focused primarily on related work that predicts the future states of objects, because we are particularly interested in developing models that may support future work towards object-based agents.

\section{Experiments and Results} \label{sec:results}

In this section we (1) demonstrate OAT's performance in a 3D room environment \cite{Hill2020Environmental, hill2020human} and compare to the current state-of-the-art object-based transition model, OP3, \cite{veerapaneni2019entity} (Section \ref{sec:results:playroom}), (2) investigate the benefit of alignment and object-level losses when training transition models (Section \ref{sec:matters}), (3) apply OAT in a real world robotics environment and show it accurately predicts both the motion of the robotic arm and its physical interaction with objects (Section \ref{sec:robotics}).

\subsection{Objects-Align-Transition Results in a 3D Room.} \label{sec:results:playroom}

\begin{figure}[h]
    \centering
    \includegraphics[width=\columnwidth]{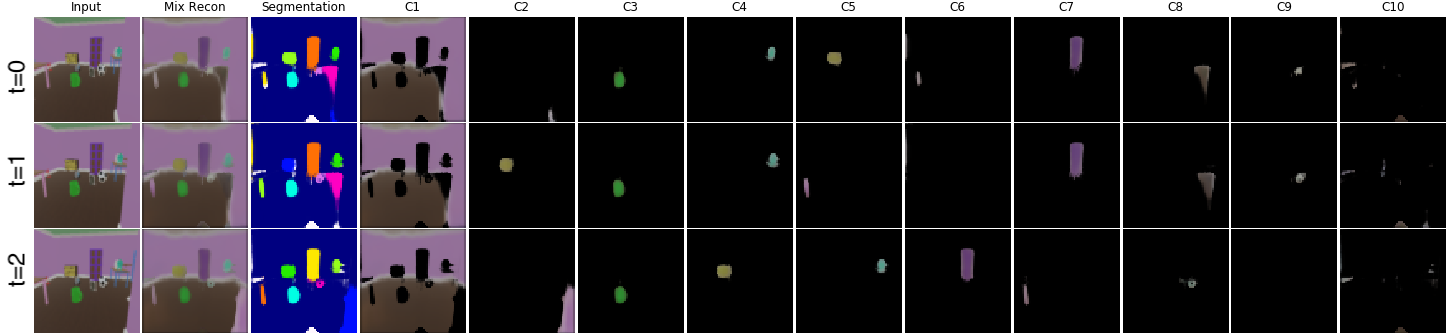}
    \caption{\textbf{Outputs of our scene decomposition and representation (MONet) module} across time. MONet predicts $K=10$ latent object representation vectors $\{z_{t,i}\}_{i=1,...,K}$ at each time-step. Here we visualise those vectors, in columns C1 to C10, using MONet's decoder. Notice that objects switch slots across time. This makes it difficult to (a) compute losses between predictions and targets and (b) integrate information about an object across time.}
    \label{fig:playroom_monet}
\end{figure}

We train and test our model, OAT, on data collected by an agent taking actions in a 3D room environment \cite{Hill2020Environmental, hill2020human}. Each procedurally generated room in the dataset contains between 10 and 45 objects from 34 different classes in 10 different colours and three different sizes. A dataset of observation-action trajectories, $(x_t, a_t)_{t=0, 1, ..., 20}$, is generated by an agent taking actions according to a learned policy in a procedurally generated room. We collect $100,000$ trajectories with a $7:2:1$ train:validation:test split. The first column of Figure \ref{fig:playroom_monet} shows an example trajectory.

We train OAT with four encoding steps (i.e. the model sees four observations for the first four time-steps) and six unrolling steps. MONet outputs $K$ object representations, $\{z_{t,i}\}_{i=1,...,K}$. We use $K=10$ objects, with $F=32$ features, and a memory with $M=12$ slots. Figure \ref{fig:playroom_monet} visualises the MONet outputs for the first $10$ steps. Our model achieves good segmentation (see the Adjusted Rand Index, ARI, in Table \ref{tab:compare} for segmentation metrics). Importantly, notice that the slots are not stable across time. For example, the yellow object in slot C5 at $t=0$ switches slots at $t=\{1, 2\}$.

\begin{figure}
    \centering
    \includegraphics[width=\columnwidth]{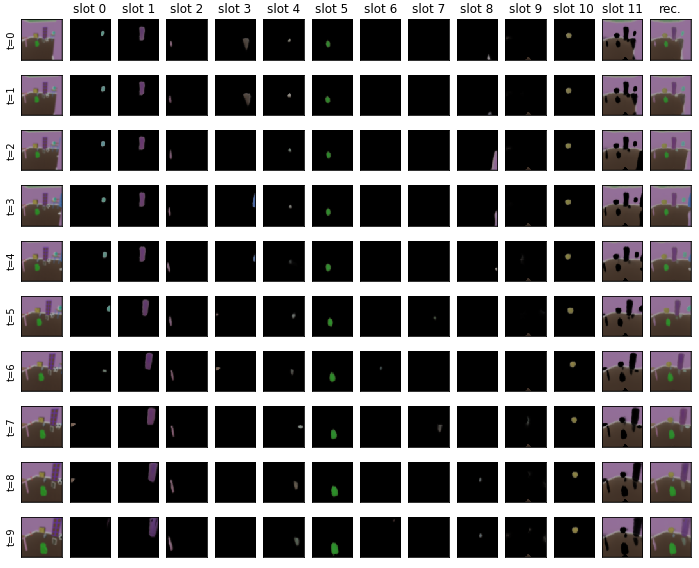}
    \caption{\textbf{Aligned inputs (first 4 time-steps) and targets (last 6 time-steps).} Our alignment module outputs object representation vectors which we visualise here using MONet's entity decoder. Most of the objects are now in consistent slots across time, making it easier to compute semantically meaningful losses directly between object representations using a simple L2 loss. Refer to Figure \ref{fig:enc} to see how the losses are computed. Notice that while MONet outputs 10 slots, the AlignNet has 12 slots.}
    \label{fig:playroom_align}
\end{figure}

The object representations, $z_t \in \Re ^{[K \times F]}$, output by MONet are fed to the alignment module. The outputs of the alignment module, $z^a_t \in \Re ^{[M \times F]}$, are visualised in Figure \ref{fig:playroom_align}. The outputs of the alignment module are object representations, Figure \ref{fig:playroom_align} is a visualisation of these representations using MONet's decoder. Our alignment module successfully keeps objects in consistent slots across time. 

\begin{figure}[h]
    \centering
    \includegraphics[width=\columnwidth]{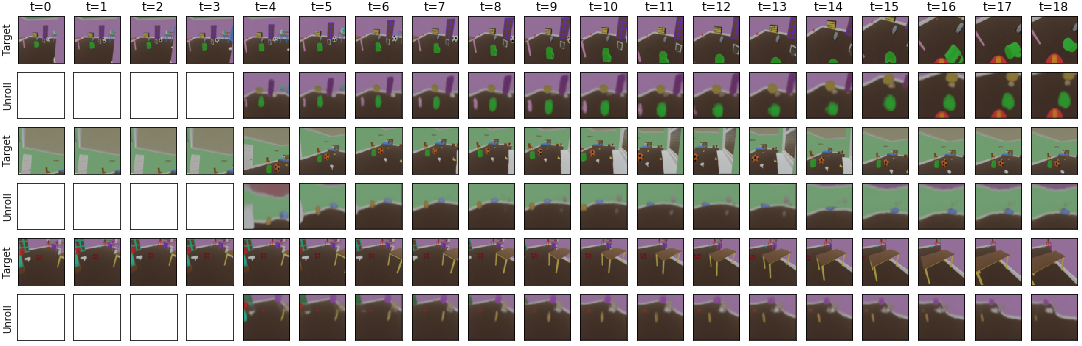}
    \caption{\textbf{Unrolling OAT for more time-steps than those seen during training.} The slot-wise transition model makes predictions in latent space (one latent per object), which we visualise using MONet's decoder. The slot-wise transition model was trained with 4 inputs steps and to unroll for 6 steps. Here the model takes the first 4 steps as input and is unrolling for 15. On the top row at $t=16$, it is impressive that the model is able predict the appearance of the avatar's base (the red circle with the yellow stripe) when looking down. (For a per-object visualisation see Figure \ref{fig:unroll_components} in the Appendix.)}
    \label{fig:playroom_unroll}
\end{figure}

The output of the alignment module is used to generate both inputs and targets for training the transition model. The transition model predicts latents, $z^d_t \in \Re^{[M \times F]}$. In Figure \ref{fig:playroom_unroll} we visualise roll-outs from our model using MONet's decoder; the reconstructed scene visualisations are generated as the mask-weighted sum of the each slot's pixels, $\sum_k \tilde{\mu}_{k, t} \tilde{x}_{t, k}$.  While the model is trained to unroll for six steps, here we unroll for 15 steps. The transition model only sees the first four frames. In the top example we see that the model is able to predict the appearance of the avatar well and in general we notice that the model given the agents actions is able to predict the position of objects well and without the representations degrading (for comparison to baselines see Figure \ref{fig:MAT_vs_baselines}). In some of the examples the targets appear to have more objects than those seen in the unroll: this is because the model has only seen the first four frames and has not seen those other objects in the room and therefore does not have enough information to predict where unseen objects will appear.

To conclude these experiments we compare our model to the current state-of-the-art object-based transition model, OP3 \cite{veerapaneni2019entity}, and to our ablated model, OT, without the alignment module in Table \ref{tab:compare} and Figure \ref{fig:MAT_vs_baselines}. Qualitatively, from Figure \ref{fig:MAT_vs_baselines} we see that OAT significantly outperforms both baselines. Unrolls using OT, trained without alignment, lead to objects merging towards a grey cloud in the middle of the frame. Unrolls from OP3 lead to objects fading into the background (we explore the cause of each of these pathologies in the next Section \ref{sec:matters}). Results were consistent for each model across multiple runs.

To quantitatively compare models we consider three metrics: \textbf{Encoding ARI}, \textbf{Unroll Pixel Error} and \textbf{Unroll ARI}. The Encoding ARI (Adjusted Rand Index, \citet{hubert1985comparing, rand1971objective}), measures the accuracy of the object segmentation masks learned by the scene decomposition model. The Unroll Pixel Error and Unroll ARI evaluate the quality of the transition models' unrolls. To compute each of these we decode the latents, $z_t^d$, predicted by unrolling the transition model, to produce images, $x_t^d$, and masks, $\mu_{k, t}^d$. The Unroll Pixel Error is the mean-squared-error between the ground truth images and $x_t^d$ for the unroll steps only. The Unroll ARI is the accuracy of the decoded masks $\mu_{k, t}^d$ compared with the ground-truth masks. For both ARI scores we exclude background pixels from the score since accurate decomposition of the objects is the main concern here. The Unroll ARI is a more meaningful evaluation of the unrolls than the Unroll Pixel Error since it is not affected by the background which often accounts for most of the pixels. Note that we only use ground-truth masks for evaluation purposes.

The results in Table \ref{tab:compare} further demonstrate the crucial role that alignment (AlignNet) plays when performing unrolls. Without alignment the Unroll ARI is significantly higher because we are not able to compute a semantically meaningful object-level loss for training. Additionally, we see that our model significantly out performs OP3 on all metrics (see Figure \ref{fig:MAT_vs_baselines}). In light of the results in Table \ref{tab:compare}, in the next section we will more concretely look at the role of alignment and the object-level loss (between object representations) when training transition models.

\begin{table}[h!]
    \centering
    \begin{tabular}{M{0.4\linewidth}|M{0.13\linewidth}M{0.13\linewidth}M{0.13\linewidth}}
         & Encoding ARI  & Unroll Pixel Error & Unroll ARI\\
         \hline
        OAT (ours) &  \textbf{0.62} & \textbf{0.0121} & \textbf{0.42}\\ 
        OT (ours, no AlignNet) &  0.59 & 0.0143  & 0.12\\ \hline
        OP3 &  0.32 & 0.0132 &  0.33\\ \hline
    \end{tabular}
    \caption{\textbf{Comparing our model OAT to baselines.} We trained three OAT and OT models and report the ARI score, Unroll Pixel error and Unroll ARI for the model with the best Unroll ARI score. Similar for OP3 except we did 10 runs since we saw more variance in the OP3 results (see Appendix \ref{sec:rep_op3} for details).}
    \label{tab:compare}
\end{table}

\subsection{What Matters For Object-centric Transition Models?}\label{sec:matters}

In this section we demonstrate the need for both (1) \textbf{alignment}, which ensures that each slot in the transition model receives the same object consistently across time and (2) \textbf{object-level loss}, computed between predicted and target object representations, rather than a pixel-level loss. We also compare different transition model cores and find a transformer followed by a slot-wise LSTM to be best.

In this section (Section \ref{sec:matters}) only, we use a MONet model that is trained using ground-truth masks instead of learning its own masks. We do this here to directly analyse the benefits of aligned object representations and object-level losses for the transition model without other confounds. Using ground-truth masks allows us to know each objects true identity across time and to directly measure the role of alignment in transition models without confounding errors from AlignNet or MONet's segmentation quality. We use a pre-trained MONet (with fixed weights) to compute object representations to keep the model similar to the full end-to-end model described in the rest of the paper. 

We train slot-wise transition models under four different conditions; feeding aligned or unaligned object representations (and target) object representations to the transition model\footnote{for the unaligned inputs we shuffle the order of the ground-truth masks $\in \Re^{[K, W, H, 3]}$ that are fed to MONet along the $K$ axis, for aligned inputs we ensure that each slot contains a consistent object across time.} and train the model with a pixel-level or object-level loss. Results in Table  \ref{tab:role_of_alignment} clearly demonstrate the benefits of (1) using aligned object representations for training transition models and (2) training transition models with a object-level loss. These results are critical for future development of transition models and demonstrates the need for alignment in transition models. Note that for the experiments using the unaligned inputs we computed the object-level loss using the Hungarian algorithm which is a lower bound estimate of the true object-level loss (since the minimum assignment in $L_2$ may not be the correct assignment).

\begin{figure}[h]
    \centering
    \includegraphics[width=0.5\textwidth]{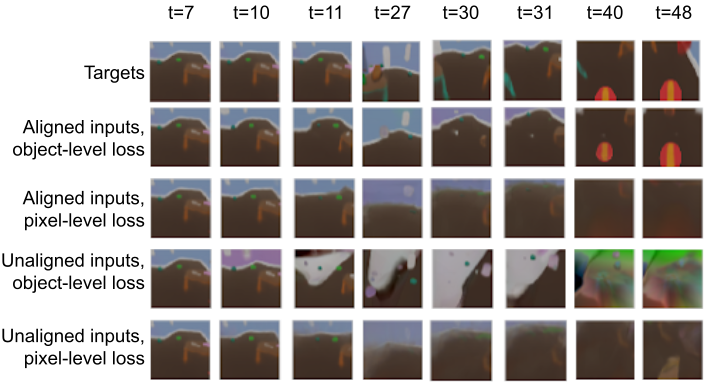}
    \caption{\textbf{Comparing slot-wise transition models trained with aligned vs. unaligned input latents and pixel-level vs. object-level loss.} These models were trained with 8 inputs steps and 12 unroll steps. Here they are being unrolled for 40 time-steps\footnote{We were able to train with more time-steps than we did when training OAT because we did not have to run MONet's large segmentation network.} The output of the transition model is a slot-wise object-based representation for each time-step. Here, we visualise the object representation vectors using MONet's decoder.  Only models trained with aligned inputs were able to predict the reappearance of the chair at $t=30$. Only the model trained with aligned inputs and object-level loss is able to predict the appearance of the avatars "base". In models trained with pixel-level loss predictions becomes very blurred.}
    \label{fig:role_of_alignment}
\end{figure}

Additionally, Figure \ref{fig:role_of_alignment} compares models trained under the four conditions listed above, we see that models trained using aligned inputs (and targets) are able to predict re-appearance of objects while those without aligned latents are not (Figures \ref{fig:role_of_alignment}, $t=30$). We also see that models trained with an object-level loss and without alignment deteriorate quickly. Additionally, for models trained with a pixel-level loss we notice a "ghosting" effect where objects fade into the background across time, similar to the effects seen in the OP3 results (Figure \ref{fig:op3_rollouts}). Pixel-level loss leads to this "ghosting" effect because most pixels in the observation are background pixels and so the background pixels dominate the loss.

\begin{table}[h]
    \centering
    \begin{tabular}{c|c c}
    \multicolumn{3}{c}{Reporting object-level error for models trained using:}\\
    & Object-level loss & Pixel-level loss \\
    \hline
    Aligned inputs & \textbf{0.0982} & 1.04 \\
    Unaligned inputs & $\geq$ 19.9 & $\geq$ 229 \\
\end{tabular}
    \caption{\textbf{The role of alignment and object-level loss when training transition models.} This table reports the latent error between predicted and ground truth latents. For models trained using unaligned input latents we compute the object-level loss using the Hungarian algorithm which is a lower bound on the actual value.}
    \label{tab:role_of_alignment}
\end{table}

We found these results to be consistent across multiple runs (see Figure \ref{fig:ordered_vs_unordered} in the Appendix) and different choices for slot-wise transition model architectures. For the results shown in this section we used a transformer with a slot-wise LSTM. Figure \ref{fig:transition_cores}, in the Appendix, compares object and pixel errors for different transition module cores. Training with object-level loss and aligned inputs, the transformer with slot-wise LSTM worked best.

\subsection{Application to Robotics}\label{sec:robotics}

Here we demonstrate the application of OAT to a real world dataset \cite{cabi2019scaling} of robot trajectories. These trajectories involve a robot arm interacting with three coloured objects of varying shapes and colours. The dataset is particularly challenging because our model must learn to predict not just the motion of the robot arm, given the arm actions, but it must also learn to predict how the arm interacts with objects requiring some understanding of intuitive physics.

We train OAT with four input steps and to unroll for six steps. Our model achieves excellent segmentation results, shown in Figure \ref{fig:robot_seg} of the Appendix. Figure \ref{fig:robot_unroll} shows impressive results obtained when unrolling our model, OAT, for significantly more steps than those seen during training. What is more in the 3rd sample from the top, we see that the model correctly predicts the reappearance of the red objects after it had been fully occluded at $t=3$ for $51$ time-steps, reappearing fully at $t=54$ in both the prediction and the target frame. Our model, OAT, is able to accurately predict the reappearance of objects, even through longer term occlusion, because the model explicitly models the history of each object endowing the model with a notion of object persistence.

\begin{figure}
    \centering
    \includegraphics[width=\columnwidth]{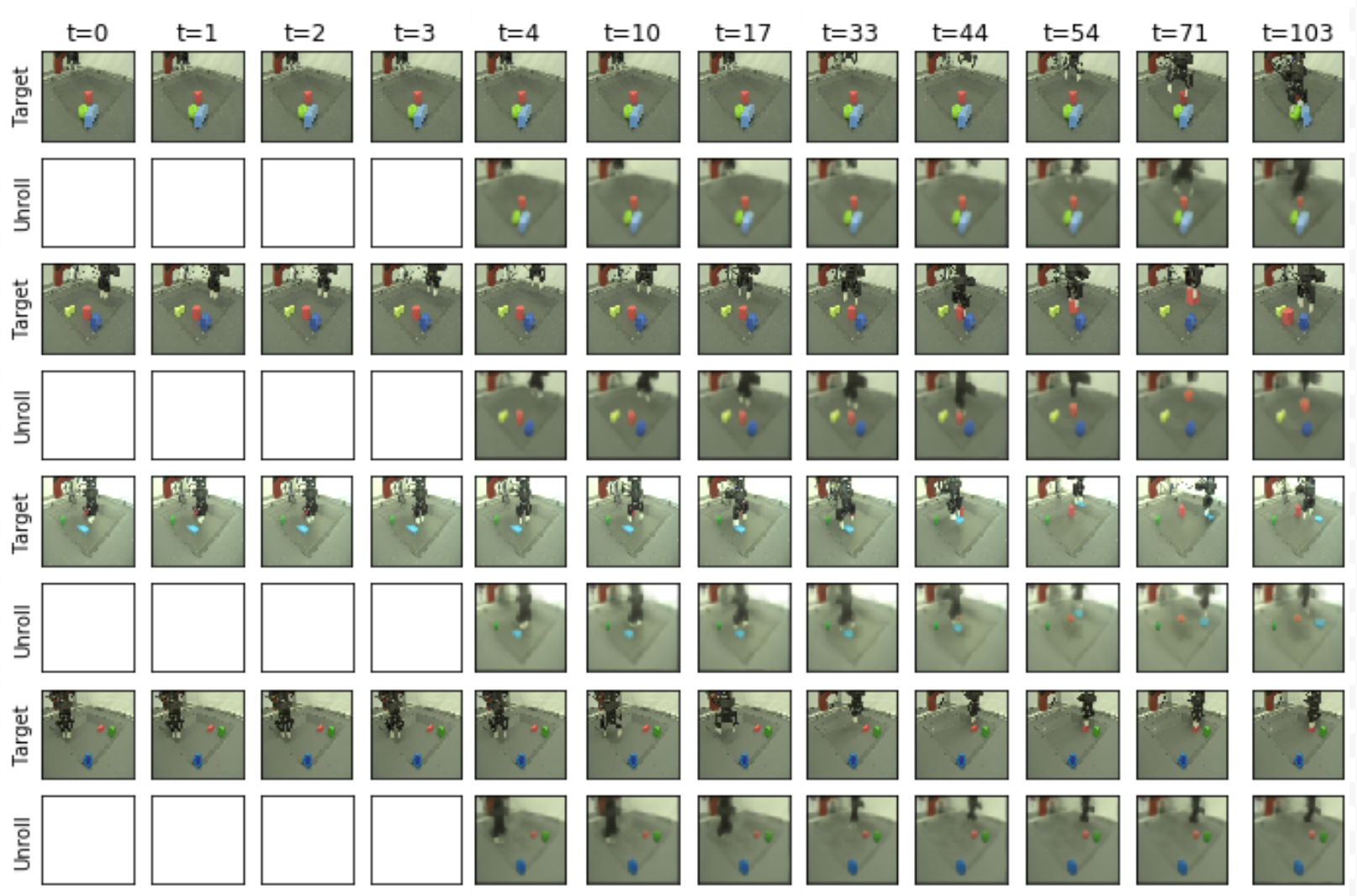}
    \caption{\textbf{Unrolling for many more steps than those seen during training.} OAT is trained to take four input steps and to unroll for six time-steps, here we demonstrate the model unrolling for 100 time-steps. Impressively, we see that our model learns both to predict the behaviour of the robotics arm accurately and how the arm interacts with the objects. We also see that the model is able to unroll for significantly more steps that those seen during training. Note that our model makes prediction in latent space; here we are visualising those latents using MONet's decoder.}
    \label{fig:robot_unroll}
\end{figure}

\section{Conclusion}

We presented Objects-Align-Transition (OAT), an object-centric transition model that combines a scene decomposition and object representation module (MONet) with a slot-wise transition module, via an alignment module. The alignment module plays two key roles. Firstly, it ensures that the slot-wise transition model receives slot-consistent object representations across time. Secondly, it allows us to compute an object-level loss rather than a pixel-level loss which is commonly used when training transition models. 

In an ablation study, we demonstrated the essential role that alignment and object-level losses play when training transition models. Additionally, we significantly out performs existing state-of-the-art object-centric transition models in a 3D partially observable environment, and we applied our model to a real-world robotics dataset, predicting many steps further into the future than those seen during training.

There is room to improve our model in the future, for example by making stochastic predictions about the future, and by better modelling the uncertainty about the objects in the environment. Meanwhile, our work paves the way for future object-centric agent research, for example, enabling agents to plan over future trajectories in object representation space.

\subsubsection*{Acknowledgments}
We would like to thank Kyriacos Nikiforou, Marta Garnelo and Adam Kosiorek for their help revising this paper and Daniel Zoran for their help with implementing baselines.

\bibliography{arxiv2021}

\begin{thebibliography}{36}
\providecommand{\natexlab}[1]{#1}
\providecommand{\url}[1]{\texttt{#1}}
\expandafter\ifx\csname urlstyle\endcsname\relax
  \providecommand{\doi}[1]{doi: #1}\else
  \providecommand{\doi}{doi: \begingroup \urlstyle{rm}\Url}\fi

\bibitem[Baillargeon et~al.(1985)Baillargeon, Spelke, and
  Wasserman]{baillargeon1985object}
Baillargeon, R., Spelke, E.~S., and Wasserman, S.
\newblock Object permanence in five-month-old infants.
\newblock \emph{Cognition}, 20\penalty0 (3):\penalty0 191--208, 1985.

\bibitem[Burgess et~al.(2019)Burgess, Matthey, Watters, Kabra, Higgins,
  Botvinick, and Lerchner]{burgess2019monet}
Burgess, C.~P., Matthey, L., Watters, N., Kabra, R., Higgins, I., Botvinick,
  M., and Lerchner, A.
\newblock {MON}et: Unsupervised scene decomposition and representation.
\newblock \emph{arXiv preprint arXiv:1901.11390}, 2019.

\bibitem[Cabi et~al.(2019)Cabi, G{\'o}mez~Colmenarejo, Novikov, Konyushkova,
  Reed, Jeong, Zolna, Aytar, Budden, Vecerik, et~al.]{cabi2019scaling}
Cabi, S., G{\'o}mez~Colmenarejo, S., Novikov, A., Konyushkova, K., Reed, S.,
  Jeong, R., Zolna, K., Aytar, Y., Budden, D., Vecerik, M., et~al.
\newblock Scaling data-driven robotics with reward sketching and batch
  reinforcement learning.
\newblock \emph{arXiv}, pp.\  arXiv--1909, 2019.
\newblock URL \url{https://sites.google.com/corp/view/data-driven-robotics}.

\bibitem[Co-Reyes et~al.(2019)Co-Reyes, Veerapaneni, Chang, Janner, Finn, Wu,
  Tenenbaum, and Levine]{reyes2019learning}
Co-Reyes, J.~D., Veerapaneni, R., Chang, M., Janner, M., Finn, C., Wu, J.,
  Tenenbaum, J., and Levine, S.
\newblock Discovering, predicting, and planning with objects.
\newblock \emph{International Conference on Machine Learning Workshop}, 2019.

\bibitem[Creswell et~al.(2020)Creswell, Nikiforou, Vinyals, Saraiva, Kabra,
  Matthey, Burgess, Reynolds, Tanburn, Garnelo, et~al.]{creswell2020alignnet}
Creswell, A., Nikiforou, K., Vinyals, O., Saraiva, A., Kabra, R., Matthey, L.,
  Burgess, C., Reynolds, M., Tanburn, R., Garnelo, M., et~al.
\newblock Alignnet: Unsupervised entity alignment.
\newblock \emph{arXiv preprint arXiv:2007.08973}, 2020.

\bibitem[Dai et~al.(2019)Dai, Yang, Yang, Carbonell, Le, and
  Salakhutdinov]{dai2019transformer}
Dai, Z., Yang, Z., Yang, Y., Carbonell, J., Le, Q.~V., and Salakhutdinov, R.
\newblock Transformer-xl: Attentive language models beyond a fixed-length
  context.
\newblock \emph{arXiv preprint arXiv:1901.02860}, 2019.

\bibitem[Garnelo \& Shanahan(2019)Garnelo and Shanahan]{garnelo2019reconciling}
Garnelo, M. and Shanahan, M.
\newblock Reconciling deep learning with symbolic artificial intelligence:
  representing objects and relations.
\newblock \emph{Current Opinion in Behavioral Sciences}, 29:\penalty0 17--23,
  2019.

\bibitem[Garnelo et~al.(2016)Garnelo, Arulkumaran, and
  Shanahan]{garnelo2016towards}
Garnelo, M., Arulkumaran, K., and Shanahan, M.
\newblock Towards deep symbolic reinforcement learning.
\newblock \emph{arXiv preprint arXiv:1609.05518}, 2016.

\bibitem[Graves et~al.(2008)Graves, Liwicki, Fern{\'a}ndez, Bertolami, Bunke,
  and Schmidhuber]{graves2008novel}
Graves, A., Liwicki, M., Fern{\'a}ndez, S., Bertolami, R., Bunke, H., and
  Schmidhuber, J.
\newblock A novel connectionist system for unconstrained handwriting
  recognition.
\newblock \emph{IEEE transactions on pattern analysis and machine
  intelligence}, 31\penalty0 (5):\penalty0 855--868, 2008.

\bibitem[Greff et~al.(2016)Greff, Rasmus, Berglund, Hao, Valpola, and
  Schmidhuber]{greff2016tagger}
Greff, K., Rasmus, A., Berglund, M., Hao, T., Valpola, H., and Schmidhuber, J.
\newblock Tagger: Deep unsupervised perceptual grouping.
\newblock In \emph{Advances in Neural Information Processing Systems}, pp.\
  4484--4492, 2016.

\bibitem[Greff et~al.(2019)Greff, Kaufman, Kabra, Watters, Burgess, Zoran,
  Matthey, Botvinick, and Lerchner]{greff2019multi}
Greff, K., Kaufman, R.~L., Kabra, R., Watters, N., Burgess, C., Zoran, D.,
  Matthey, L., Botvinick, M., and Lerchner, A.
\newblock Multi-object representation learning with iterative variational
  inference.
\newblock In \emph{International Conference on Machine Learning}, pp.\
  2424--2433. PMLR, 2019.

\bibitem[Ha \& Schmidhuber(2018)Ha and Schmidhuber]{ha2018world}
Ha, D. and Schmidhuber, J.
\newblock World models.
\newblock \emph{arXiv preprint arXiv:1803.10122}, 2018.

\bibitem[Hafner et~al.(2019)Hafner, Lillicrap, Ba, and
  Norouzi]{hafner2019dream}
Hafner, D., Lillicrap, T., Ba, J., and Norouzi, M.
\newblock Dream to control: Learning behaviors by latent imagination.
\newblock \emph{arXiv preprint arXiv:1912.01603}, 2019.

\bibitem[He et~al.(2018)He, Li, Liu, He, and Barber]{he2018tracking}
He, Z., Li, J., Liu, D., He, H., and Barber, D.
\newblock Tracking by animation: Unsupervised learning of multi-object
  attentive trackers.
\newblock \emph{arXiv preprint arXiv:1809.03137}, 2018.

\bibitem[Hill et~al.(2020{\natexlab{a}})Hill, Lampinen, Schneider, Clark,
  Botvinick, McClelland, and Santoro]{Hill2020Environmental}
Hill, F., Lampinen, A., Schneider, R., Clark, S., Botvinick, M., McClelland,
  J.~L., and Santoro, A.
\newblock Environmental drivers of systematicity and generalization in a
  situated agent.
\newblock In \emph{International Conference on Learning Representations},
  2020{\natexlab{a}}.
\newblock URL \url{https://openreview.net/forum?id=SklGryBtwr}.

\bibitem[Hill et~al.(2020{\natexlab{b}})Hill, Mokra, Wong, and
  Harley]{hill2020human}
Hill, F., Mokra, S., Wong, N., and Harley, T.
\newblock Human instruction-following with deep reinforcement learning via
  transfer-learning from text.
\newblock \emph{arXiv preprint arXiv:2005.09382}, 2020{\natexlab{b}}.

\bibitem[Hsieh et~al.(2018)Hsieh, Liu, Huang, Fei-Fei, and
  Niebles]{hsieh2018learning}
Hsieh, J.-T., Liu, B., Huang, D.-A., Fei-Fei, L.~F., and Niebles, J.~C.
\newblock Learning to decompose and disentangle representations for video
  prediction.
\newblock In \emph{Advances in Neural Information Processing Systems}, pp.\
  517--526, 2018.

\bibitem[Hubert \& Arabie(1985)Hubert and Arabie]{hubert1985comparing}
Hubert, L. and Arabie, P.
\newblock Comparing partitions.
\newblock \emph{Journal of classification}, 2\penalty0 (1):\penalty0 193--218,
  1985.

\bibitem[Kipf et~al.(2019)Kipf, van~der Pol, and Welling]{kipf2019contrastive}
Kipf, T., van~der Pol, E., and Welling, M.
\newblock Contrastive learning of structured world models.
\newblock \emph{arXiv preprint arXiv:1911.12247}, 2019.

\bibitem[Kosiorek et~al.(2018)Kosiorek, Kim, Teh, and
  Posner]{kosiorek2018sequential}
Kosiorek, A., Kim, H., Teh, Y.~W., and Posner, I.
\newblock Sequential attend, infer, repeat: Generative modelling of moving
  objects.
\newblock In \emph{Advances in Neural Information Processing Systems}, pp.\
  8606--8616, 2018.

\bibitem[Lake et~al.(2017)Lake, Ullman, Tenenbaum, and
  Gershman]{lake2017building}
Lake, B.~M., Ullman, T.~D., Tenenbaum, J.~B., and Gershman, S.~J.
\newblock Building machines that learn and think like people.
\newblock \emph{Behavioral and Brain Sciences}, 40, 2017.

\bibitem[L{\"o}we et~al.(2020)L{\"o}we, Greff, Jonschkowski, Dosovitskiy, and
  Kipf]{lowe2020learning}
L{\"o}we, S., Greff, K., Jonschkowski, R., Dosovitskiy, A., and Kipf, T.
\newblock Learning object-centric video models by contrasting sets.
\newblock \emph{arXiv preprint arXiv:2011.10287}, 2020.

\bibitem[Marcus \& Davis(2019)Marcus and Davis]{marcus2019rebooting}
Marcus, G. and Davis, E.
\newblock \emph{Rebooting AI: Building Artificial Intelligence We Can Trust}.
\newblock Ballantine Books Inc., 2019.

\bibitem[Pylyshyn(1989)]{Pylyshyn1989TheRO}
Pylyshyn, Z.~W.
\newblock The role of location indexes in spatial perception: A sketch of the
  finst spatial-index model.
\newblock \emph{Cognition}, 32:\penalty0 65--97, 1989.

\bibitem[Racani\`{e}re et~al.(2017)Racani\`{e}re, Weber, Reichert, Buesing,
  Guez, Jimenez~Rezende, Puigdom\`{e}nech~Badia, Vinyals, Heess, Li, Pascanu,
  Battaglia, Hassabis, Silver, and Wierstra]{racaniere2017imagination}
Racani\`{e}re, S., Weber, T., Reichert, D., Buesing, L., Guez, A.,
  Jimenez~Rezende, D., Puigdom\`{e}nech~Badia, A., Vinyals, O., Heess, N., Li,
  Y., Pascanu, R., Battaglia, P., Hassabis, D., Silver, D., and Wierstra, D.
\newblock Imagination-augmented agents for deep reinforcement learning.
\newblock In \emph{Advances in Neural Information Processing Systems 30}, pp.\
  5690--5701, 2017.

\bibitem[Rand(1971)]{rand1971objective}
Rand, W.~M.
\newblock Objective criteria for the evaluation of clustering methods.
\newblock \emph{Journal of the American Statistical association}, 66\penalty0
  (336):\penalty0 846--850, 1971.

\bibitem[Ronneberger et~al.(2015)Ronneberger, Fischer, and
  Brox]{ronneberger2015u}
Ronneberger, O., Fischer, P., and Brox, T.
\newblock U-net: Convolutional networks for biomedical image segmentation.
\newblock In \emph{International Conference on Medical image computing and
  computer-assisted intervention}, pp.\  234--241. Springer, 2015.

\bibitem[Shanahan et~al.(2020)Shanahan, Beyret, Crosby, and
  Cheke]{shanahan2020artificial}
Shanahan, M., Beyret, B., Crosby, M., and Cheke, L.
\newblock Artificial intelligence and the common sense of animals.
\newblock \emph{Trends in Cognitive Sciences}, 24\penalty0 (11):\penalty0
  862--872, 2020.

\bibitem[Silver et~al.(2016)Silver, Huang, Maddison, Guez, Sifre, Van
  Den~Driessche, Schrittwieser, Antonoglou, Panneershelvam, Lanctot,
  et~al.]{silver2016mastering}
Silver, D., Huang, A., Maddison, C.~J., Guez, A., Sifre, L., Van Den~Driessche,
  G., Schrittwieser, J., Antonoglou, I., Panneershelvam, V., Lanctot, M.,
  et~al.
\newblock Mastering the game of {G}o with deep neural networks and tree search.
\newblock \emph{Nature}, 529\penalty0 (7587):\penalty0 484--489, 2016.

\bibitem[van Steenkiste et~al.(2018)van Steenkiste, Chang, Greff, and
  Schmidhuber]{van2018relational}
van Steenkiste, S., Chang, M., Greff, K., and Schmidhuber, J.
\newblock Relational neural expectation maximization: Unsupervised discovery of
  objects and their interactions.
\newblock \emph{arXiv preprint arXiv:1802.10353}, 2018.

\bibitem[Vaswani et~al.(2017)Vaswani, Shazeer, Parmar, Uszkoreit, Jones, Gomez,
  Kaiser, and Polosukhin]{vaswani2017attention}
Vaswani, A., Shazeer, N., Parmar, N., Uszkoreit, J., Jones, L., Gomez, A.~N.,
  Kaiser, {\L}., and Polosukhin, I.
\newblock Attention is all you need.
\newblock In \emph{Advances in neural information processing systems}, pp.\
  5998--6008, 2017.

\bibitem[Veerapaneni et~al.(2019)Veerapaneni, Co-Reyes, Chang, Janner, Finn,
  Wu, Tenenbaum, and Levine]{veerapaneni2019entity}
Veerapaneni, R., Co-Reyes, J.~D., Chang, M., Janner, M., Finn, C., Wu, J.,
  Tenenbaum, J.~B., and Levine, S.
\newblock Entity abstraction in visual model-based reinforcement learning,
  2019.

\bibitem[Wang et~al.(2018)Wang, Girshick, Gupta, and He]{wang2018non}
Wang, X., Girshick, R., Gupta, A., and He, K.
\newblock Non-local neural networks.
\newblock In \emph{Proceedings of the IEEE Conference on Computer Vision and
  Pattern Recognition}, pp.\  7794--7803, 2018.

\bibitem[Watters et~al.(2019)Watters, Matthey, Bosnjak, Burgess, and
  Lerchner]{watters2019cobra}
Watters, N., Matthey, L., Bosnjak, M., Burgess, C.~P., and Lerchner, A.
\newblock Cobra: Data-efficient model-based rl through unsupervised object
  discovery and curiosity-driven exploration, 2019.

\bibitem[Weis et~al.(2020)Weis, Chitta, Sharma, Brendel, Bethge, Geiger, and
  Ecker]{weis2020unmasking}
Weis, M.~A., Chitta, K., Sharma, Y., Brendel, W., Bethge, M., Geiger, A., and
  Ecker, A.~S.
\newblock Unmasking the inductive biases of unsupervised object representations
  for video sequences, 2020.

\bibitem[Zaheer et~al.(2017)Zaheer, Kottur, Ravanbakhsh, Poczos, Salakhutdinov,
  and Smola]{zaheer2017deep}
Zaheer, M., Kottur, S., Ravanbakhsh, S., Poczos, B., Salakhutdinov, R.~R., and
  Smola, A.~J.
\newblock Deep sets.
\newblock In \emph{Advances in neural information processing systems}, pp.\
  3391--3401, 2017.

\end{thebibliography}
\bibliographystyle{arxiv2021}

\clearpage

\appendix

\section*{Appendix}

\section{Model Details}\label{sec:details}

\subsection{Objects-Align-Transition Implementation Details}\label{sec:OAT_details}

MONet and AlignNet each have their own set of hyper-parameters. OAT introduces only one additional parameter, $\zeta$, which weights the transition model loss. While we used $\zeta=10$ for all the results in this paper, we found that the model performed similarly well for smaller values of $\zeta$, such as $\zeta=1.0$. For MONet, we use the same hyper-parameters as those detailed in the Appendix 1.B of \cite{burgess2019monet}, except we use $\gamma=0.05$ for the mask KL weight rather than $\gamma=0.5$. Qualitatively, our model still performed well with $\gamma=0.5$. We train OAT for 2 million steps (30 days, though the model appears to have converged after 1M steps) with an effective batch size of 32 (batch size of four spread over eight NVIDIA V100 GPUs) and a learning rate of $3\times 10^{-4}$. We have to use a small batch size because MONet is processing a whole sequence of images and this requires more memory (for the activations) than processing a single image.

For the results on the robotics dataset we used the same hyper-parameters as above, except we use $\gamma=0.5$. The model is trained for 1.6 million steps (16 days, however the model is close to convergence after 200k steps) using $K=7$ object slots, $M=8$ memory slots and a MONet feature size, $F=32$.

\subsection{Additional Details and Results for the Ablation.}

Recall that in the ablation study \textbf{only}, we train MONet using ground-truth masks. This means that rather than using an attention network (U-net) to predict masks, $\mu_{k,t}$, we use ground-truth masks and feed these to the slot-wise VAE.

\begin{figure}[h!]
    \centering
    \includegraphics[width=0.5\textwidth]{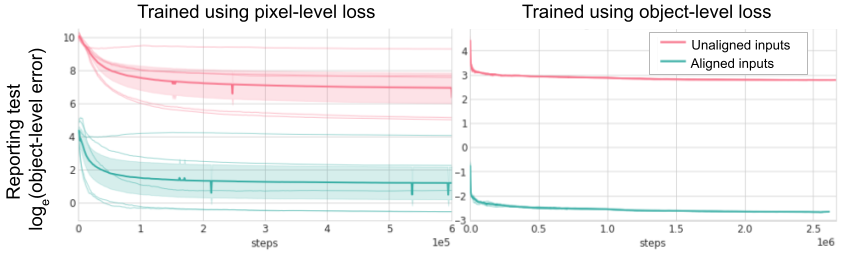}
    \caption{\textbf{Comparing models trained using aligned vs. unaligned inputs and pixel-level vs object-level loss.} We see that, across 5 runs, training transition models using aligned inputs and latent-level loss obtains best results.}
    \label{fig:ordered_vs_unordered}
\end{figure}

Figure \ref{fig:ordered_vs_unordered} compares models trained using aligned vs. unaligned inputs and pixel-level vs. object-level loss, showing 5 runs per model configuration. Models trained using latent loss can do $10\times$ the number of updates per second compared to models trained using the pixel loss.

\begin{figure}
    \centering
    \includegraphics[width=0.5\textwidth]{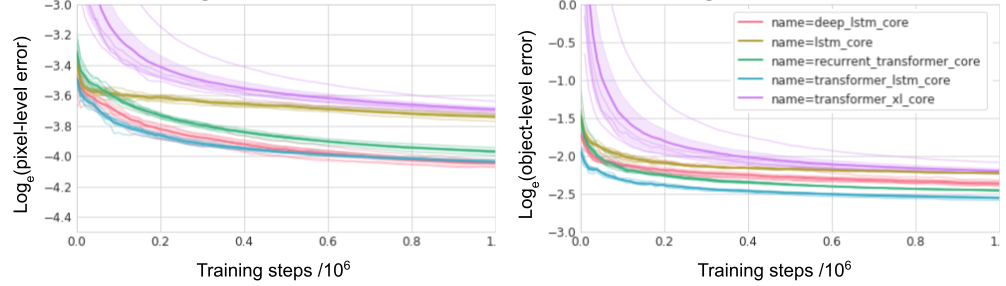}
    \caption{\textbf{Comparing transition module cores.} We show five runs for each model configuration. We find that a transformer with slot LSTM outperforms the Transformer XL \cite{dai2019transformer}, as well as LSTMs \cite{graves2008novel},  Deep LSTMs and a recurrent transformer that predicts both a state and outputs at each time-step and appends that state to the input to make predictions at the next time-step.}
    \label{fig:transition_cores}
\end{figure}

Figure \ref{fig:transition_cores} shows the effect of using different slot-wise transition model cores. We see that a transformer followed by a slot-wise LSTM achieves the best results when training using an object-level loss and aligned inputs.

\section{Additional Results on the 3D Room Dataset.}

\begin{figure}
    \centering
    \includegraphics[width=0.5\textwidth]{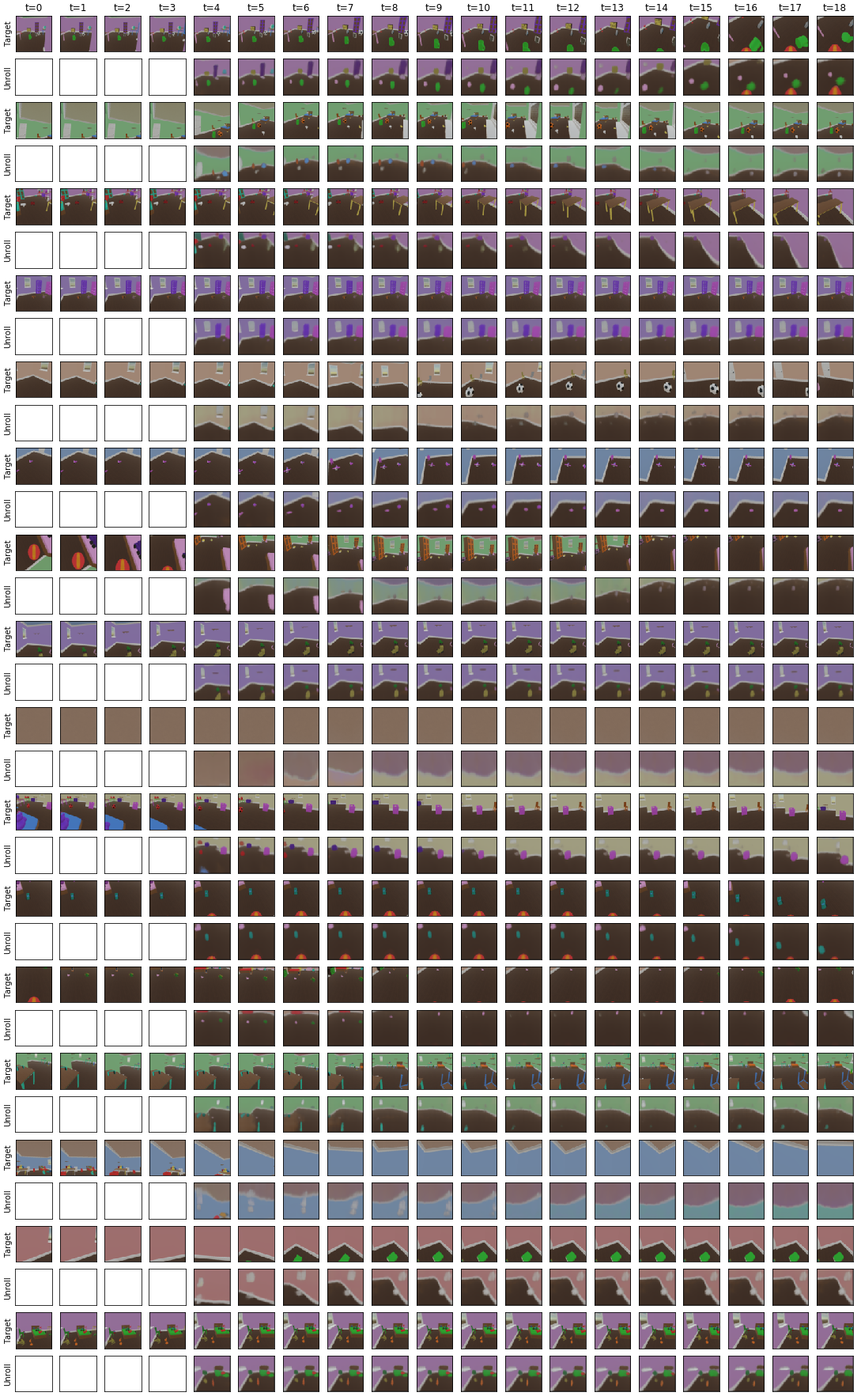}
    \caption{\textbf{Additional OAT unrolls.} OAT was trained to take four input steps and unroll for six time-steps, here we demonstrate OAT unrolling for 15 time-steps. Note, this model was trained using the exact same training set-up (detailed in Section \ref{sec:OAT_details} in the Appendix) as results shown in Figure \ref{fig:playroom_unroll} but is a different seed. We see that we get similarly good results across multiple seeds.}
    \label{fig:additional_mat_unroll}
\end{figure}

Figure \ref{fig:additional_mat_unroll} shows additional unrolls using OAT, trained under the same conditions as those described in Section \ref{sec:OAT_details}. The figure shows consistently good results across many samples.

\begin{figure}
    \centering
    \includegraphics[width=0.5\textwidth]{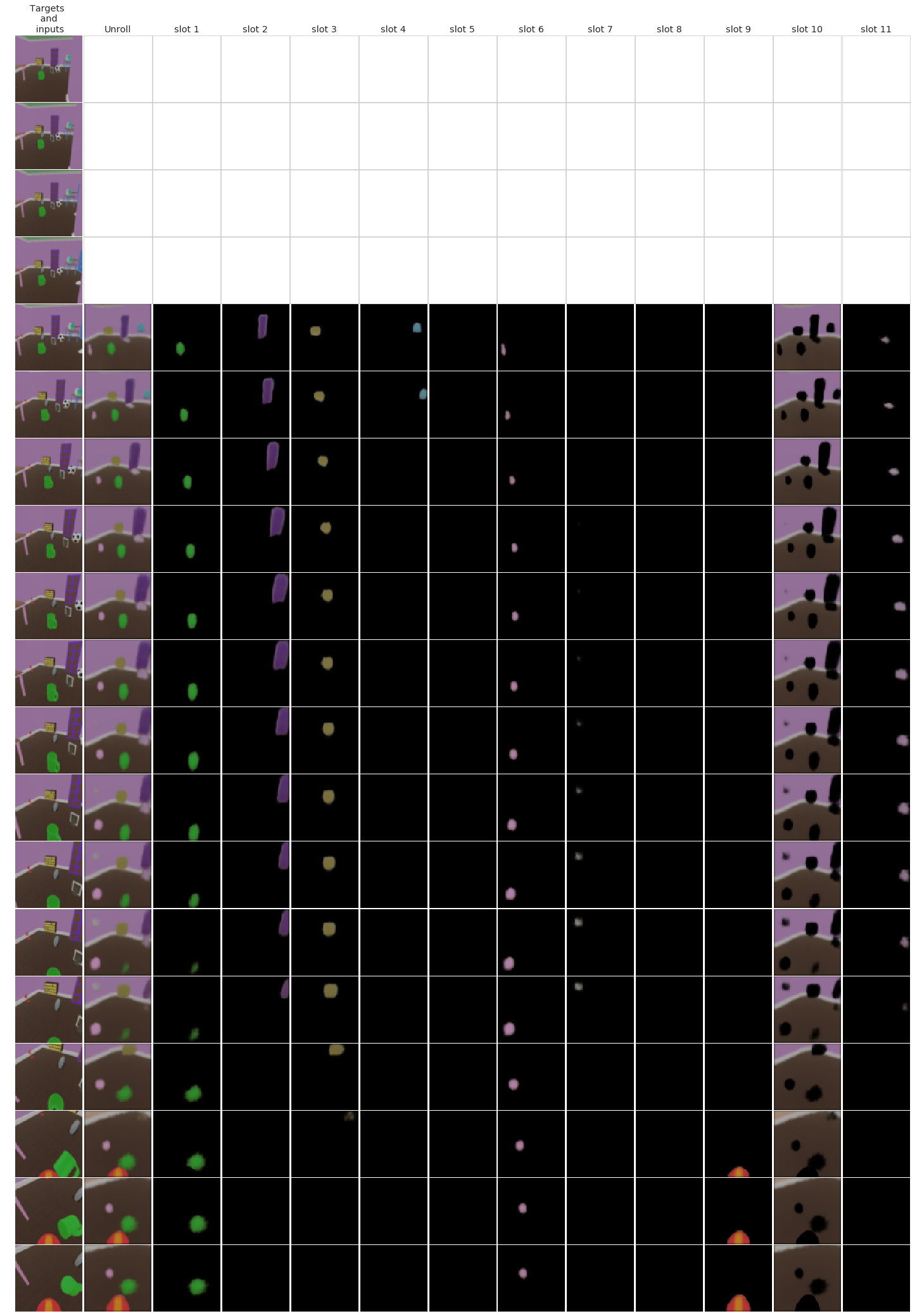}
    \caption{\textbf{Showing the components predicted during the unrolls.} Rows correspond to time-steps. Slot one to 11 show the objects predicted during the unroll. OAT sees the first four time-steps and unrolls for the next 15 time-steps.}
    \label{fig:unroll_components}
\end{figure}

In the main text we visualise OAT's unrolls by passing each predicted object representation to MONet's decoder function to reconstruct objects, $\tilde{x}_{t,k}$ and masks $\tilde{\mu}_{t, k}$, and combining these into a single image for each time-step, $\sum_{k=1}^K \tilde{\mu}_{t, k} \tilde{x}_{t,k}$. In Figure \ref{fig:unroll_components} rather than combining the component objects, we show each of the reconstructed, masked objects, $\tilde{x}_{t, k} \tilde{\mu}_{t, k}$ across time.

\section{Additional Results on the Robotics Dataset.}

\begin{figure}
    \centering
    \includegraphics[width=\columnwidth]{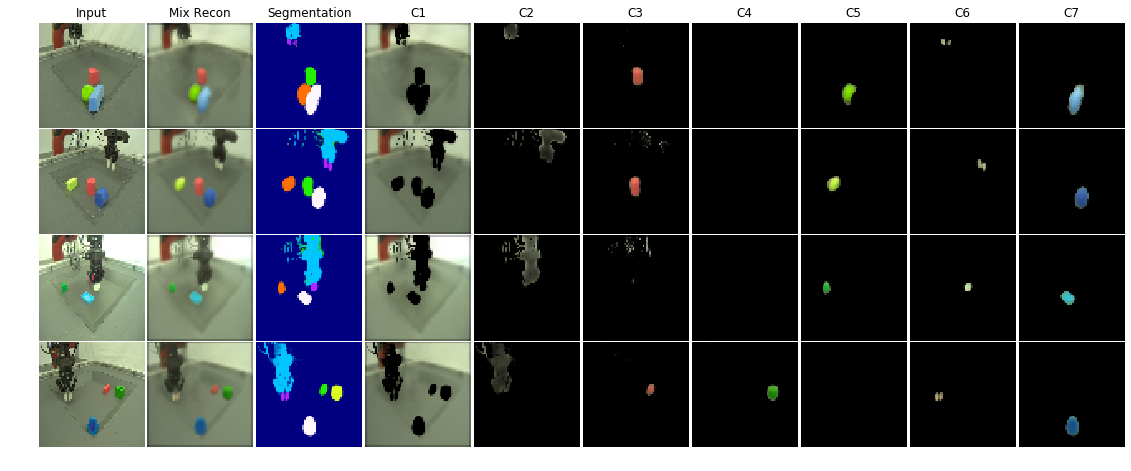}
    \caption{\textbf{Scene decomposition and representation module outputs for the real world robotics data.} Each object is correctly placed in its own column.}
    \label{fig:robot_seg}
\end{figure}

Figure \ref{fig:robot_seg} visualises the MONet outputs on the Robotics dataset. We see that the arm, gripper and objects are each represented in their own object slot, C1 to C7.

\section{Reproducing OP3 Results}\label{sec:rep_op3}

OP3 is a slot-wise scene segmentation and dynamics model. It applies a refinement network (based on IODINE, \citet{greff2019multi}) to an initial estimate of the scene's slot-wise object features, followed by an action-conditioned prediction model to predict the features at the next time step. The refinement and prediction steps are interlaced through a sequence of steps, making OP3 the closest baseline to OAT.

\begin{figure}[t]
    \centering
    \includegraphics[width=0.5\textwidth]{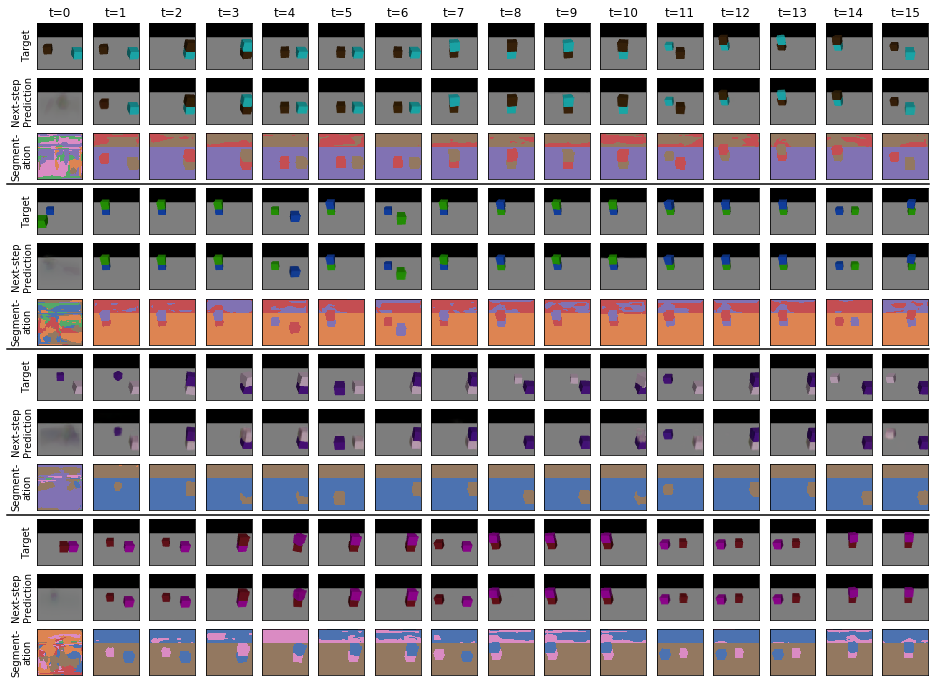}
    \caption{OP3 \cite{veerapaneni2019entity} results on the \texttt{pickplace\_multienv\_10k} dataset released by the OP3 authors. This model was trained as per OP3's standard regime with one dynamics step and one refinement step at each time-step (i.e. no unroll). We've plotted the decoded outputs post-dynamics but pre-refinement at each time-step to show the strength of the transition model. Though the segmentations are imperfect (partly because we used 7 component slots), and the model occasionally drops an object (e.g. sequence 3), it has learnt the dataset's jumpy object transitions perfectly.}
    \label{fig:op3_dataset}
\end{figure}

We first tested our implementation on the \texttt{pickplace\_multienv\_10k} dataset made available by the authors of OP3. We trained and evaluated the model using one refinement step and a next-step prediction in a loop. The results in Figure \ref{fig:op3_dataset} show that our implementation has learnt the dynamics well enough to cope with the dataset’s jumpy object transitions. 

\begin{figure}
    \centering
    \includegraphics[trim={0 0 0 1.4cm},clip,width=0.5\textwidth]{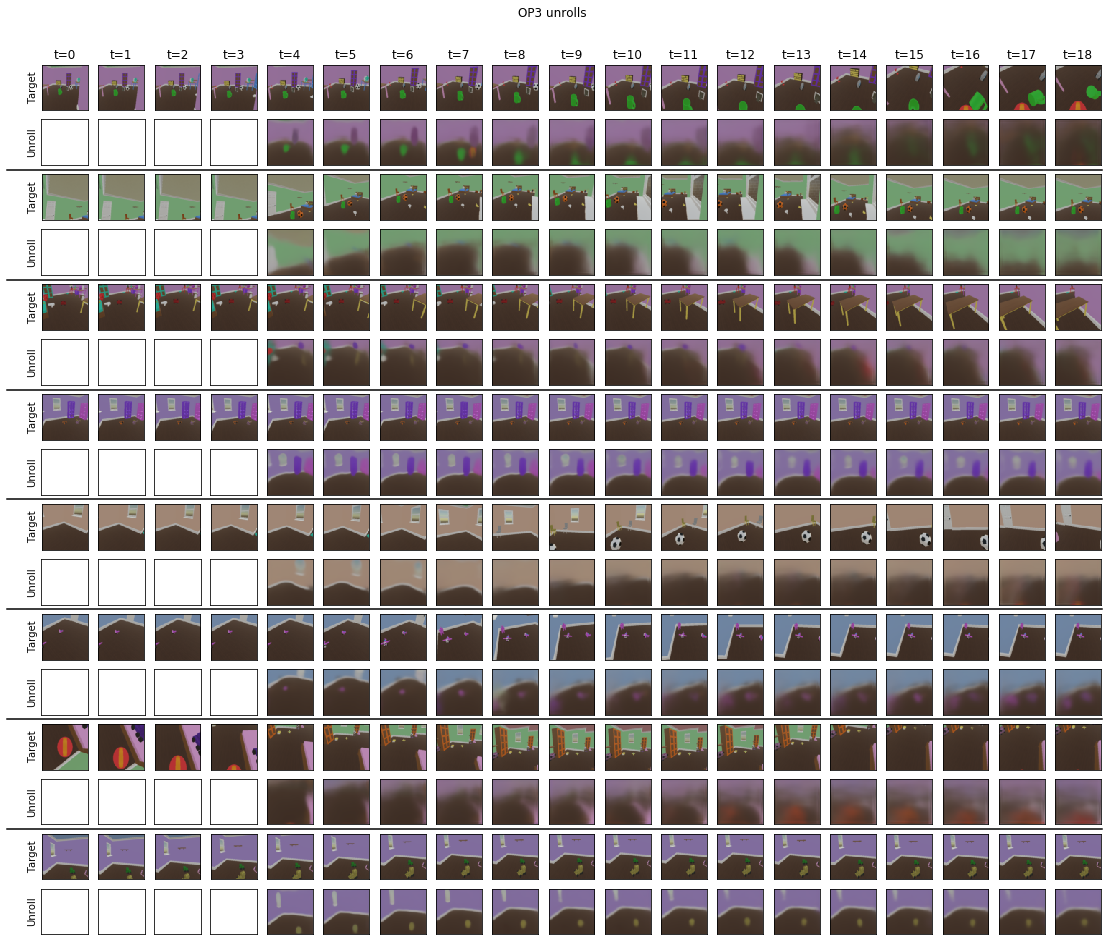}
    \caption{\textbf{Baseline OP3 \cite{veerapaneni2019entity} rollouts.} This model was trained with the same number of input steps (four) and unroll steps (six) as our model, OAT (see Figure \ref{fig:playroom_unroll}). Here we picked the model with the best ARI score from 10 independent runs. The results are noteworth at t=4 (following the burn-in refinement steps) for accurately placing objects. But for t>4, the prediction core quickly begins to accumulate errors, distorting the size and position of objects and in some cases the floor edges. It further fails to predict the appearance of the avatar (for instance, toward the end of the first sequence), which is an easy-to-learn and predictable consequence of the action space.}
    \label{fig:op3_rollouts}
\end{figure}

Following this validation of our implementation, we trained OP3 on the Playroom dataset in a regime that mirrors ours. Four burn-in steps with refinement and next-step prediction allow the model to build its initial estimate of the scene. These are followed by six unroll steps during training, where the refinement is disabled. (Here the slot parameters are only updated via the prediction core.) For evaluation, we roll-out for 15 steps instead of six to test the dynamics model at long-range prediction. Figure \ref{fig:op3_rollouts} shows eight such roll-out sequences from the model with the best ARI score over the rollout steps.

\subsubsection{OP3 Hyperparameters}

We substituted OP3's original relation net-based transition model with a transformer module plus SlotLSTM. This is identical to OAT's setup and allows the fairest comparison. The transformer uses 2 layers, 4 heads, and embedding size 128. The SlotLSTM's hidden size is also 128.

We train OP3 with the refinement and prediction schedule described above, again similar to OAT. The refinement encoder is a convolutional network with five layers with [64, 128, 128, 256, 256] output channels, kernel shape 5, and stride 1, followed by an MLP with [256, 256] hidden units. The slots have 64 latents each (and hence the MLP outputs 128 posterior parameters). We use stochastic latents only, avoiding OP3's deterministic latents without loss of generality.

The encoder, as in IODINE, is applied slotwise across the following refinement inputs: the input image, the log likelihood of the image with respect to the predicted output distribution, the current estimate of the slot parameters, the gradient of the log likelihood with respect to the slot parameters, the logits of the decoded object masks, the masks themselves, the gradient of the log likelihood with respect to the masks, and a counterfactual (as in \citet{greff2019multi}).

The decoder is a broadcast decoder comprising transpose convolutions with [64, 64, 64, 64, 64, 4] output channels, kernel size 5, and stride 1. The decoded mask logits are activated with a tanh scaled by 10.0.

We used a KL loss scale of 0.5 and a fixed output distribution scale of 0.1. To stabilize training, we also clipped gradients to a norm of 5.0. For the refinement inputs, we clipped gradients to a slightly higher norm of 10.0. Finally, we used an effective batch size of 32, the RMSProp optimizer, and a learning rate of 1e-5. We trained all OP3 models for 4 million steps (15 days).

\end{document}